\documentclass[letterpaper, 10 pt, conference]{ieeeconf}  % Comment this line out if you need a4paper

\IEEEoverridecommandlockouts                              % This command is only needed if 
\usepackage{graphics}
\usepackage{epsfig}
\usepackage{mathptmx}
\usepackage{times}
\usepackage{amsmath}
\usepackage{amssymb}
\usepackage{hyperref}
\usepackage{float}
\usepackage{graphicx}
\usepackage{multirow}
\usepackage{array}
\usepackage[inkscapelatex=false]{svg}
\usepackage{subfigure}
\usepackage{subcaption}
\usepackage{xcolor}
\newcommand{\ph}[1]{{\textbf{#1:}}}
\usepackage[font=small,skip=1pt]{caption}

\title{\LARGE \bf
Pixels-to-Graph: Real-time Integration of Building Information Models and Scene Graphs for Semantic-Geometric Human-Robot Understanding~\looseness=-1

}

\author{
Antonello Longo$^{1,2}$, 
Chanyoung Chung$^{3}$, 
Matteo Palieri$^{3}$,
Sung-Kyun Kim$^{3}$,\\
Ali Agha$^{3}$,
Cataldo Guaragnella$^{2}$ and
Shehryar Khattak$^{1}$
\thanks{$^{1}$ NASA Jet Propulsion Laboratory, California Institute of Technology, Pasadena, CA, USA.}
\thanks{$^{2}$ Department of Electrical And Information Engineering, Polytechnic University of Bari, Italy.}
\thanks{$^{3}$ Field AI, Mission Viejo CA, USA. (Work conducted at NASA Jet Propulsion Laboratory).}
\thanks{The research was carried out at the Jet Propulsion Laboratory, California Institute of Technology, under a contract with the National Aeronautics and Space Administration (80NM0018D0004).}
\thanks{\copyright 2025. All rights reserved.}
}

\begin{document}

\maketitle
\thispagestyle{empty}
\pagestyle{empty}

\begin{abstract}
%The human-machine interaction paradigm is rapidly evolving as a consequence of recent developments in the field of Natural Language Processing (NLP). Machines are now able to interpret the semantics expressed in human-like input commands. However, how to build high level, semantic-rich representations of complex environments to be used as a mental model during navigation is still an open challenge. 
Autonomous robots are increasingly playing key roles as support platforms for human operators in high-risk, dangerous applications. To accomplish challenging tasks, an efficient human-robot cooperation and understanding is required. While typically robotic planning leverages 3D geometric information, human operators are accustomed to a high-level compact representation of the environment, like top-down 2D maps representing the Building Information Model (BIM). 3D scene graphs have emerged as a powerful tool to bridge the gap between human readable 2D BIM and the robot 3D maps. In this work, we introduce Pixels-to-Graph (Pix2G), a novel lightweight method to generate structured scene graphs from image pixels and LiDAR maps in real-time for the autonomous exploration of unknown environments on resource-constrained robot platforms. To satisfy onboard compute constraints, the framework is designed to perform all operation on CPU only.
The method output are a de-noised 2D top-down environment map and a structure-segmented 3D pointcloud which are seamlessly connected using a multi-layer graph abstracting information from object-level up to the building-level.
% A 2D structural segmentation model trained on the CubiCasa5K dataset is proposed and a novel GAN-based approach to deal with map noise is also introduced. 
The proposed method is quantitatively and qualitatively evaluated during real-world experiments performed using the NASA JPL NeBula-Spot legged robot to autonomously explore and map cluttered garage and urban office like environments in real-time.~\looseness=-1

% Index terms - Robot perception, 3D scene graphs, Scene understanding, Structural segmentation, Generative Adversarial Networks
\end{abstract}

%%%%%%%%%%%%%%%%%%%%%%%%%%%%%%%%%%%%%%%%%%%%%%
% Introuction
\section{INTRODUCTION}
\label{sec:intro}
%Autonomous mobile robots are gradually gaining importance in our daily activities, playing a fundamental role in the development of applications that are no longer limited to the industrial field but are increasingly approaching people's everyday needs, \cite{wong2018autonomous}. %The term autonomy represents not only the capability of taking decisions based on personal reasoning, without external intervention, but also the ability of managing uncertainty.

%Autonomous mobile robots are increasingly approaching people's everyday needs. They are asked to robustly carry out complex tasks while being deployed in dynamic environments. To allow human supervision during robotic operations, autonomous agents are also required to use the same semantics as human operators   \cite{hendrikx2021connecting} \cite{chen2023automated}. Moreover, the human-like abstract and high-level representation of concepts, enables a faster and optimized problem solving strategy. The human knowledge must be transferred to robots in order to successfully accomplish those tasks. Building Information Maps (BIMs) emerged as robust means to bridge the gap between human representation and robotic operation models. BIMs allow encoding buildings' geometry data (spatial and structural) along with semantic information. The introduction of semantic knowledge from human concepts like types of objects and rooms, enables advanced reasoning capabilities for efficient navigation and human-robot interaction \cite{ruiz2017building}.

Autonomous mobile robots are increasingly utilized for augmenting human actions in everyday operations. Given their maturing abilities to robustly carry out complex tasks in dynamic and challenging environments, they are especially being deployed in dirty and dangerous applications where the risk to human lives is high. Nevertheless, in applications like infrastructure inspection and disaster response, robotic autonomy still needs human operator support for carrying out the complex decision making process. The decision making process is typically guided by the situational awareness provided by the robot and transmitted to human operators: detailed and time-critical situational awareness provision leads to more accurate and efficient mission strategies. There is therefore a critical need to encode the information gathered by autonomous agents in a more human understandable representation in real-time~\cite{hendrikx2021connecting} \cite{chen2023automated}.

\begin{figure}[ht!]
    \centering
    \includegraphics[width=\columnwidth]{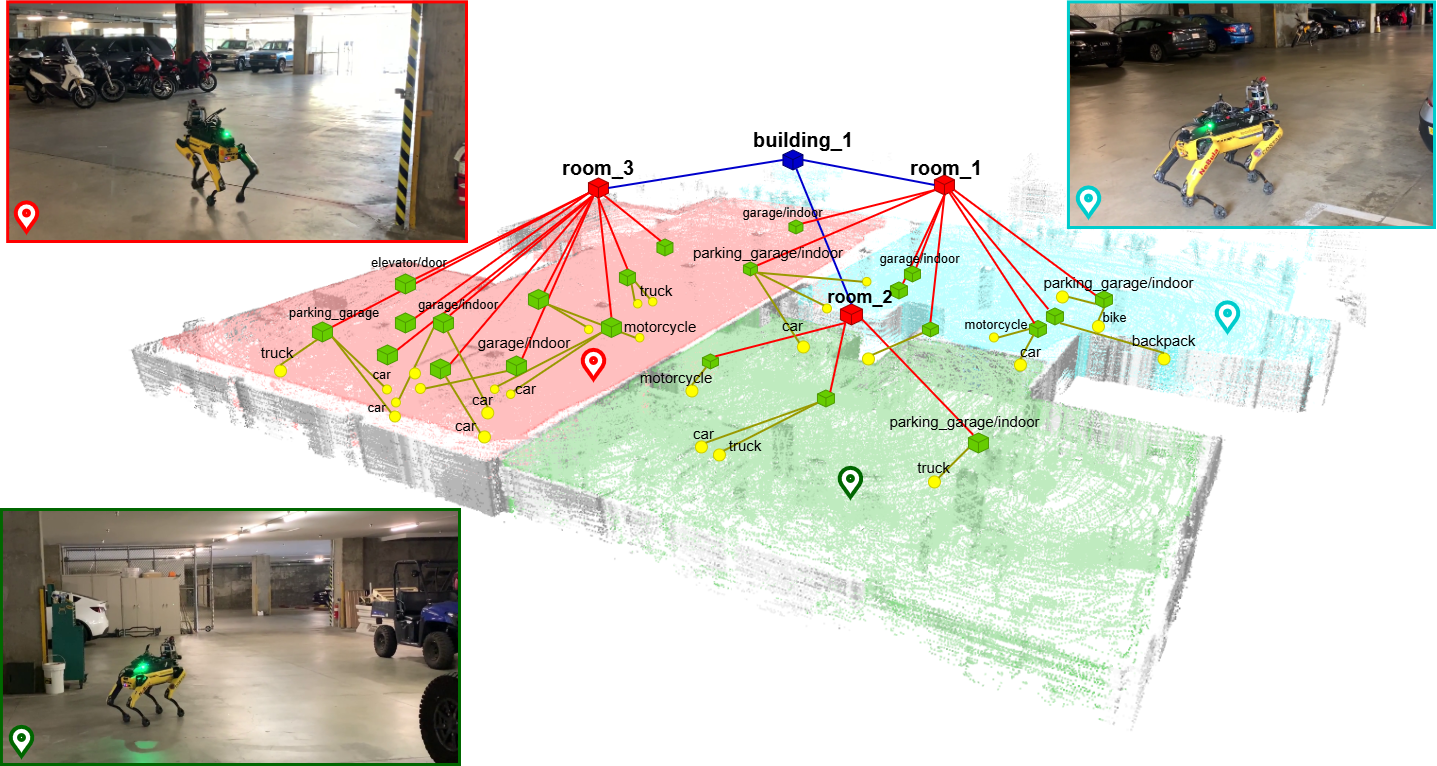}
    \caption{3D scene graph with human-readable semantic BIM generated on board the NeBula-Spot robot during autonomous operations.}
    \label{fig:concept}
\end{figure}

While robots formulate action plans based on detailed 3D maps with geometric and semantic information, human operators' reasoning usually requires a higher level understanding of the whole environment. 2D Building Information Modeling (BIM) maps are typically used to jointly represent buildings' geometric data along with semantic knowledge from human concepts like types of objects and rooms in a compact form for quick understanding~\cite{ruiz2017building}. 
Scene graphs recently emerged as a solution to bridge the gap between low-level spatial information and high-level structural and semantic information \cite{hughes2022hydra}, \cite{rosinol20203d}, \cite{Werby-RSS-24}. %, \cite{armeni20193d}. %They are multi-layer structures in which nodes are spatial concepts and edges are their relations. 
Using scene graphs as human-robot shared representation allows to achieve optimized action planning based on a hierarchical, abstract model of the environment and its action space. 
% The first layer of a scene graph usually encodes information about artifacts. Instance detection and segmentation algorithms are employed to build fine-grained object-level representations. %Thanks to the availability of more powerful machines that can run very complex models, autonomous robots can benefit from the advancements of state-of-the-art segmentation models to gather semantic information about objects spread in the surroundings. 
% At a higher level of abstraction, the different logical spaces of an environment must be identified. Structural segmentation models are typically used to break down the footprint of a big environment into independent components (e.g. rooms, corridors). 
However, sensor noise and partial acquisitions in presence of clutter highly affect the accuracy of the geometric space partitioning and thus the accuracy of 3D scene graph accuracy. In this case, generative modeling has proven to be a good solution to tackle the problem of noisy or missing information~\cite{henry2021pix2pix}. %\cite{yeh2017semantic} 
While scene graphs are good assets to enable human support during autonomous robotic operations in challenging environments, how to build such a semantically rich representation in real-time, resource-constrained scenarios still remains an open challenge.~\looseness=-1

Motivated by these challenges, this work presents Pix2G, a novel framework to generate a hierarchical scene graph that seamlessly connects an accurate 2D top-down environment map and a structure-segmented 3D pointcloud for human understanding, enabling the human operator to conduct autonomous robotic explorations of unknown environments while receiving actionable situational awareness in real-time.
% hierarchical scene graphs in real-time during autonomous robotic explorations of unknown environments.
To account for onboard compute constraints during in-field operations, the framework is designed to be lightweight and achieve high performance relying on CPU only. The structural segmentation task is therefore formulated as an instance segmentation problem in the 2D image domain. Starting from the 3D LiDAR map of the robot, a top-down ortho-projected image is generated and used to gather room and floor level understanding, as a result of space partitioning. Next, we build a 3D scene graph incorporating $i)$ low-level 3D geometric information, $ii)$ object and scene distribution and $iii)$ structural components of the mapped environment. The output graph therefore acts as a connecting layer between human readable 2D BIM and the robot 3D map. By bridging this gap, human instructions and robot actions can be conveniently mapped between the two representations, enabling seamless human-robot operations for real-world critical applications.
Finally, the challenge of map noise and incomplete sensor acquisitions is also addressed in this work. A lightweight model is introduced to remove noise and add missing structural information, thus improving the segmentation accuracy. %To satisfy the computation requirements of constrained robotic platforms, the proposed framework is designed to run on CPU-only 

The contributions of our work are as follows:
\begin{itemize}
    \item \textbf{Real-time approach}: The scene understanding and structural segmentation systems operate in 2D image domain, reducing the workload of the overall framework compared to resource intensive 3D pointcloud approaches. This makes our approach suitable for CPU-only inference.
  
    \item \textbf{Robustness to sensor noise and partial acquisitions}: 
    We propose lightweight learning-based image inpainting and space partitioning approaches with CPU-only inference, trained on a large scale dataset, to make the proposed system robust to challenging scenarios and adaptable to previously unseen scenarios.

    \item \textbf{Online and offline operation modes}: The proposed approach is presented with two operation modes. Online mode is used to perform structural segmentation tasks in real-time (\url{https://youtu.be/htXL7GMhTJ0}) with an evolving map, and the offline mode can perform full map segmentation in a single inference step at the end of the exploration of the full environment or if previous maps are available.
\end{itemize}

The rest of the paper is organized as follows: section \ref{sec:barw} provides an analysis of the background and related works, section \ref{sec:method} describes the adopted method and the proposed solutions, section \ref{sec:exp} presents all experimental results. Conclusions and future works are in section \ref{sec:conclusions}. 
%%%%%%%%%%%%%%%%%%%%%%%%%%%%%%%%%%%%%%%%%%%%%%

%%%%%%%%%%%%%%%%%%%%%%%%%%%%%%%%%%%%%%%%%%%%%%
% Related work
\section{Related Work}
\label{sec:barw}
\ph{Instance segmentation}
The instance segmentation problem can be formulated as the task of classifying pixels or points with semantic labels. In the image domain, classic approaches based on computer vision techniques leverage morphological operations \cite{chudasama2015image}, while deep-learning models have yielded a new generation of segmentation models with remarkable performance improvements \cite{9356353}. A detailed description of Region-based Convolutional Neural Networks (R-CNNs) is provided in \cite{bharati2020deep}, as well as a survey on all the derived improved architectures like Fast Region-based Convolutional Network (Fast R-CNN). %\cite{Girshick_2015_ICCV}. 
Following in the same footsteps, Faster R-CNN is introduced in  as a Region Proposal Network (RPN) that shares full-image convolutional features with the detection network to enable cost-free region proposals \cite{ren2015faster}. %In  \cite{acuna2019devil} instead, authors introduce a Semantically Thinned Edge Alignment Learning (STEAL) approach for semantically-aware mask edge refinement.

Kirillov et al. \cite{kirillov2023segment} propose a real-time segmentation model based on promptable tasks, powered by Vision Transformer (ViT) to compute image embeddings. A prompt encoder embeds sparse and dense prompts and the two sources are finally combined in a mask decoder that predicts segmentation masks. 
CLIP shows how natural language supervision, in the form of image, text pairs, can be used during pre-training to learn how to perform a wide variety of computer vision tasks, \cite{radford2021learning}.
Lang-seg exploits a text encoder, computing embeddings of descriptive input labels, together with a transformer-based image encoder, to perform semantic segmentation based on user input classes \cite{DBLP:journals/corr/abs-2201-03546}. %Motivated by the idea that the structural content of images is the most informative factor to semantic segmentation that can also be shared across domains, \cite{chang2019all} propose a Domain Invariant Structure Extraction (DISE) framework to realize image-translation across domains and enable label transfer to improve segmentation performance. 

In contrast to image segmentation techniques, Nguyen et al. \cite{6758588} survey the segmentation techniques in the 3D domain. In \cite{8885691}, a network architecture that learns 2D textural appearance and 3D structural features in a unified manner is presented. The model exploits the back-projection of 2D image features into 3D coordinates.
%%%%%%%%%%%%%%%%%%%%%%%%%%%%%%%%%%%%%%%%%%%%%%%%%%%%%%
% \begin{figure*}[t]
%     \centering
%     \subfigure[Original map]{
%         \includegraphics[width=0.3\textwidth]{Pictures/b11_ok.png}
%     }
%     \subfigure[BEV image with $0.5m$ height threshold]{
%         \includegraphics[width=0.3\textwidth]{Pictures/b11_0.png}
%     }
%     \subfigure[BEV image with $1.10m$ height threshold]{
%         \includegraphics[width=0.3\textwidth]{Pictures/b11_1.png}
%     }
%     \caption{Effects of varying thresholds on BEV images generated from LiDAR maps.}
%     \label{fig:bev_thresh}
% \end{figure*}

%%%%%%%%%%%%%%%%%%%%%%%%%%%%%%%%%%%%%%%%%%%%%%%%%%%%%%

\ph{Structural segmentation}
The structural segmentation task regards the identification of independent spaces within a complex-shaped environment. ROSE$^2$ \cite{luperto2022robust} exploits the main direction of the walls to to reconstruct the geometry of structural elements and perform structural segmentation. The proposed approach is resilient to clutter but assumes a global view of the map from which wall directions can be extracted. 
%Luperto et al. \cite{luperto2022robust} identify the structure of indoor environments from 2D occupancy maps to reconstruct the geometry of structural elements. These are then used to perform structural segmentation. By identifying the main directions of walls, a clean geometrical floor-plan-like description of the environment is always extracted, making ROSE$^2$ resilient to clutter. 
Friedman et al \cite{friedman2007voronoi} instead extract a Voronoi graph from the occupancy grid map to describe environments in terms of their spatial layout and connectivity of different places. However, the Voronoi technique struggles in case of partial views of the structural components. 
%\cite{de2014statistical} present a pipeline to automatically extract structural information from human-annotated floorplan images and partition the space even in case of missing information. 

Several approaches have been proposed in the literature to label pointclouds and extract the polygonal structure of independent components, \cite{ijgi11100530} and \cite{ntiyakunze2023segmentation}. Ambrucs et al. \cite{ambrucs2017automatic} leverage accurate and robust detection of building structural elements to reconstruct a 2D floor plan from unstructured pointclouds. Nevertheless, the increased complexity of 3D segmentation results in higher computation demands and  higher inference speed when compared to 2D models. In \cite{mahmood2021learning}, two approaches to solve the structural segmentation problem in the 2D and 3D domain are introduced. Pointer Network learns the geometry of different rooms as a sequence of points, while a 2D CNN performs wall detection and extraction using 2D histograms.

\ph{Floorplan generation}
Following the Pix2Pix \cite{henry2021pix2pix} approach, generative models can also be used to reconstruct missing information in an image. In the specific context of synthetic floorplan generation, \cite{nauata2021house} and \cite{sun2022wallplan} prove to be effective in automatically building plausible structures starting from design constraints. However these works estimate probabilistic layouts from high-level configurations. 

Methods for completing 2D metric maps by completing the missing information behind closed doors are instead introduced in \cite{luperto2022reconstruction}.
%\cite{9568786} and . 

\begin{figure*}[ht!]
    \centering
    \includegraphics[width=\linewidth]{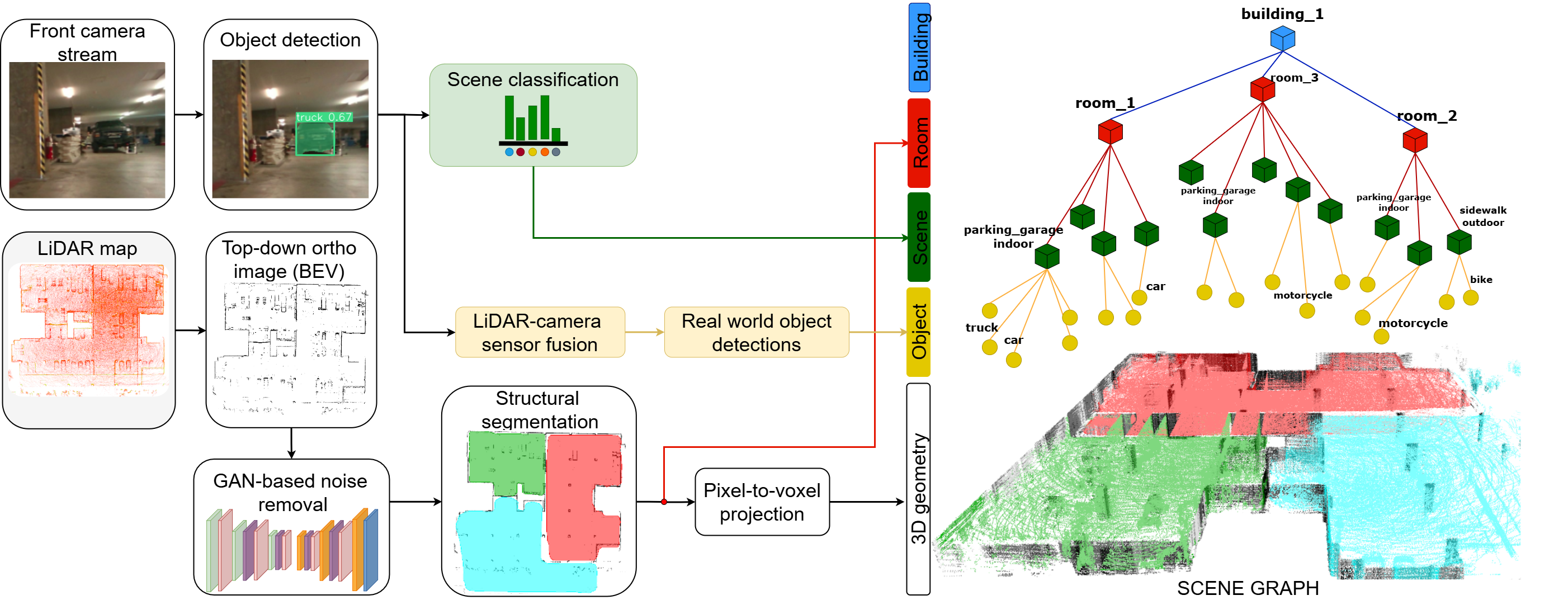}
    \caption{An overview of the proposed approach, from input data to the creation of structurally segmented 2D BIM map and the 3D scene graph. The bottom layer of the graph (object - yellow) is created using the onboard robot images. Object detections are used to capture object distributions and a LiDAR-camera sensor fusion module is responsible for projecting detected object instances to map coordinates. Object distributions are then used as an input for a scene classification model to create the second layer of the graph (scene - green). Simultaneously, a 2D Bird’s-Eye View map generation node extracts the 3D LiDAR map's noisy 2D footprint, that is first refined using a GAN-based 2D map denoising network. The denoised output map is used as an input for the 2D structure segmentation model. The output labels independent structural components in the third layer of the graph (room - red). Furthermore, the 2D structure segmentation output is also used to perform 3D pointcloud partitioning (white). From room organization, information about the building is finally obtained, represented in the top layer of the scene graph (building - blue). Video description of approach is presented in:~\url{https://youtu.be/8FPta-rKkY0}}
    \label{fig:architecture}
    \vspace{-5mm}
\end{figure*}

\ph{Semantic scene graphs}
The problem of real-time generation of hierachical scene graphs has been widely investigated in \cite{hughes2022hydra}. HYDRA is a novel framework for constructing a 3D scene graph during the exploration, with an integrated module for loop closure detection and optimization directly in the scene graphs. The engine leverages a local Euclidean Signed Distance Function (ESDF), built around the robot location, to extract topological map of places, based on Generalized Voronoi Diagram (GVD). Separate rooms are then identified from the topological map of places. However, the system lacks a scene classification module to label rooms. Moreover authors claim that their room segmentation approach, based on topological reasoning, fails to segment rooms in an open floor-plan.
In \cite{kim20193}, a 3D scene graph construction framework working on streams of images is introduced. In single image frames, Faster R-CNN recognizes object instances, while a Factorizable Net (F-Net) extracts object-to-object action, spatial and  prepositional relations.

%\subsection{GAP} 
The review of related works highlights that most of the approaches for the structural segmentation work on dense pointclouds, often assuming complete acquisitions. In resource constrained applications, this is not optimal because 3D segmentation models often have higher inference times and computation demands, \cite{zhang2022bridging}.
Methods working on 2D maps usually assume a Manhattan world distribution \cite{coughlan1999manhattan} and leverage Voronoi graphs that fail in case of incomplete structural information. Real-time systems for 3D graph generations are based on structural analysis and struggle to generate to open floorplan scenarios. 

Therefore, to the best of authors knowledge, no existing work in the literature: $i)$ generates a multi-level scene graph in challenging, large-scale environments, in presence of clutter and partial acquisitions, $ii)$ performs a refinement of missing structural information based on generative modeling and $iii)$ is built on top of lightweight segmentation models to be run onboard robotic agents in real-time. 
%%%%%%%%%%%%%%%%%%%%%%%%%%%%%%%%%%%%%%%%%%%%%%

%%%%%%%%%%%%%%%%%%%%%%%%%%%%%%%%%%%%%%%%%%%%%%
% Method
\section{METHOD}
\label{sec:method}
% %%%%%%%%%%%%%%%%%%%%%%%%%%%%%%%%%%%%%%%%%%%%%%%%%%%%%%
% \begin{figure*}[t!]
%     \centering
%     \subfigure[Original map]{
%         \includegraphics[width=0.3\textwidth]{Pictures/b11_ok.png}
%     }
%     \subfigure[BEV image with $0.5m$ height threshold]{
%         \includegraphics[width=0.3\textwidth]{Pictures/b11_0.png}
%     }
%     \subfigure[BEV image with $1.10m$ height threshold]{
%         \includegraphics[width=0.3\textwidth]{Pictures/b11_1.png}
%     }
%     \caption{Effects of varying thresholds on BEV images generated from LiDAR maps.}
%     \label{fig:bev_thresh}
% \end{figure*}
% %%%%%%%%%%%%%%%%%%%%%%%%%%%%%%%%%%%%%%%%%%%%%%%%%%%%%%
To fill-in the above mentioned gap, Pix2G solves the object detection, scene classification and structural segmentation problems in the image domain. The following sections describe the overall system architecture and then dive into the details of each sub-module. 

\ph{System architecture}
The complete workflow, as indicated in Fig. \ref{fig:architecture}, is composed of two parallel pipelines, responsible for the extraction of the structural information and semantic information respectively. The input block accesses the robot's front camera image at time $t$ and the LiDAR map built up to time $t$. 
The current frame is used to perform object detection and scene classification. The output of the two modules is used to fill the information of the first two layers of the scene graph. The second branch, in parallel utilizes the LiDAR map to generate a Bird's-Eye View (BEV) and extract a top-down orthographic projection image containing the 2D structural map footprint. This is then fed to the segmentation model that outputs independent structural components. All the information is finally merged together to render the 3D scene graph.

\subsection{Object and scene information}
To gather information about object distribution in the space under exploration, a detection framework based on MMDetection~\cite{Chen2019MMDetectionOM} is employed. For each successful detection in the image frame at time $t$ the model outputs a mask and a class ID. To register detected object's pose in the global frame, the robot pose from the LiDAR localization module is used to perform lidar-vision sensor fusion and project the object detection to 3D coordinates. Specifically, the object detection mask is projected into the 3D LiDAR map points which are then clustered using DBSCAN to get object center's 3D coordinate. Furthermore, all object detections are also used to classify the scene type using AlexNet~\cite{alexnet} trained on Places365~\cite{zhou2017places} dataset to classify the scene type as either $i)$ indoor or $ii)$ outdoor and then to determine the specific location type (e.g. kitchen, office, garage).

\subsection{Structural segmentation system}
The task is formulated as an instance segmentation problem, with the goal of identifying different independent components (rooms, corridors) and mask out each as a single instance. The structure of the environment has therefore to be expressed in a top-down image format.

\begin{figure}[t!]
    \centering
    \subfigure[Original map]{
        \includegraphics[width=0.45\columnwidth]{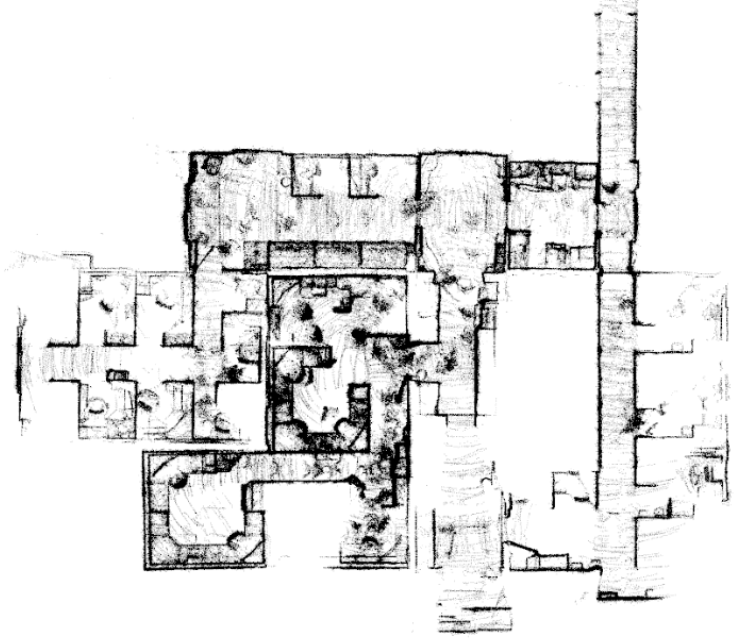}
        \label{fig:subfig1}
    }
    \subfigure[BEV image with $0.5m$ height threshold]{
        \includegraphics[width=0.45\columnwidth]{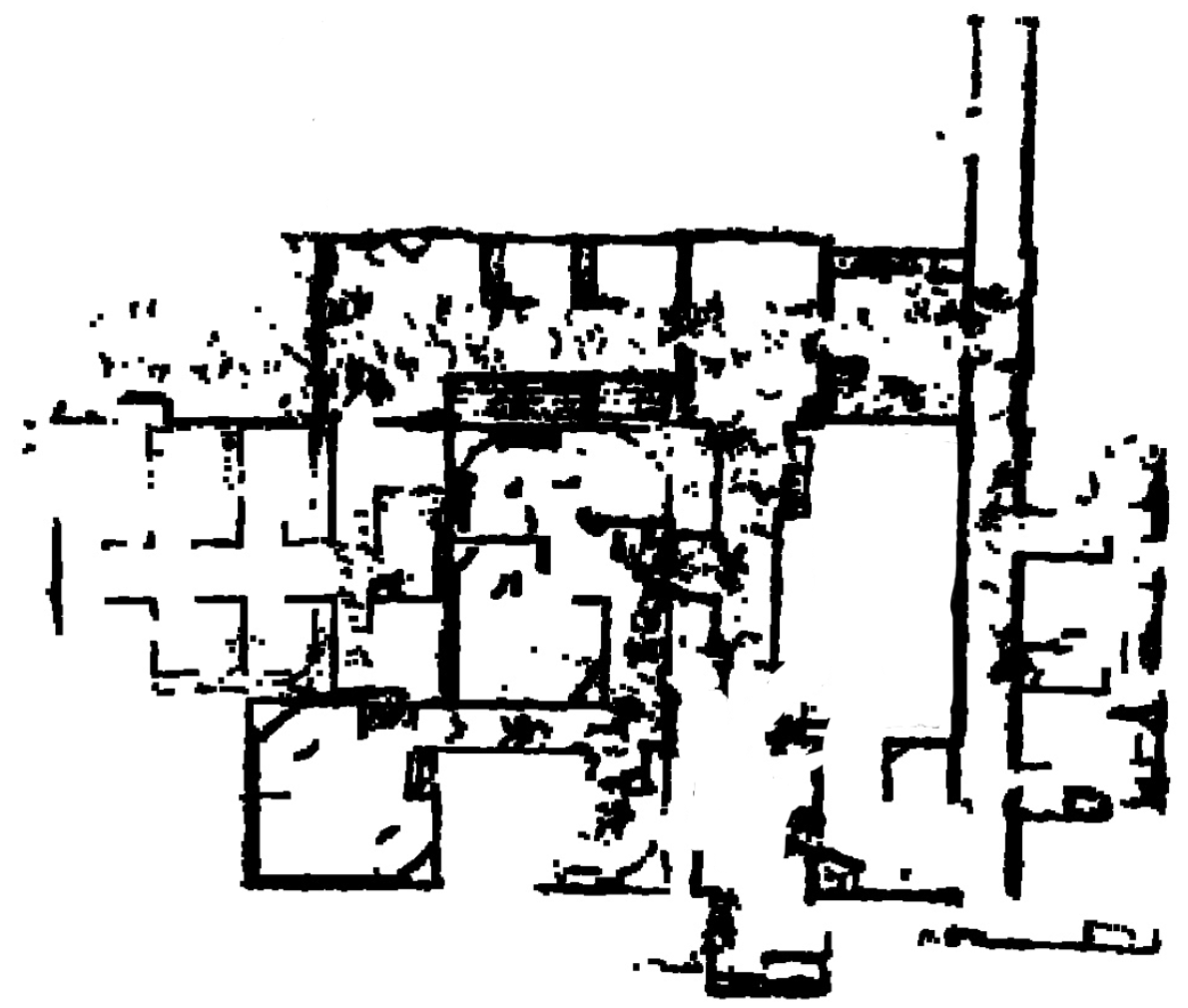}
        \label{fig:subfig2}
    }

    \subfigure[BEV image with $1.10m$ height threshold]{
        \includegraphics[width=0.45\columnwidth]{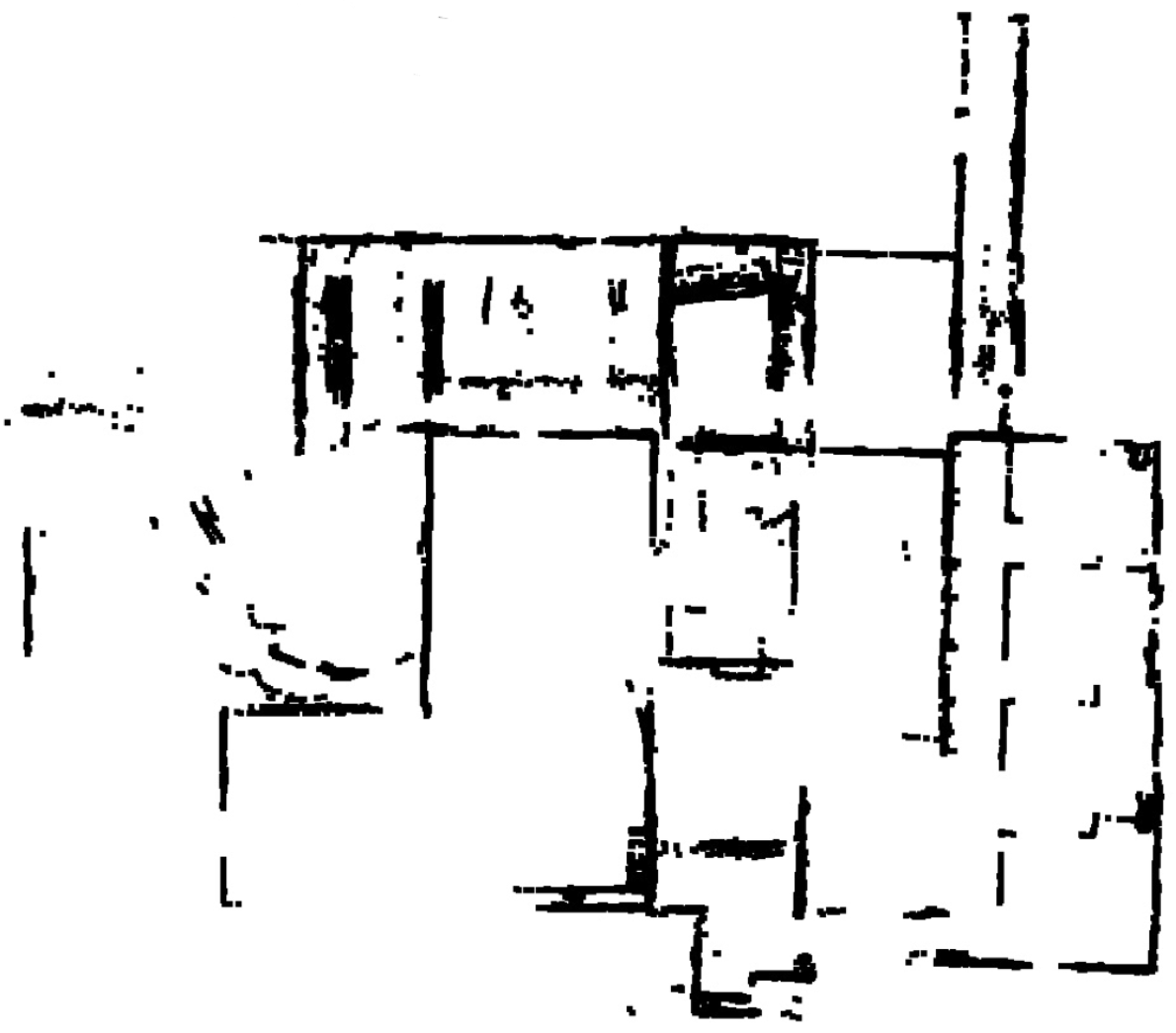}
        \label{fig:subfig3}
    }
    \subfigure[BEV image with the proposed adaptive threshold]{
        \includegraphics[width=0.45\columnwidth]{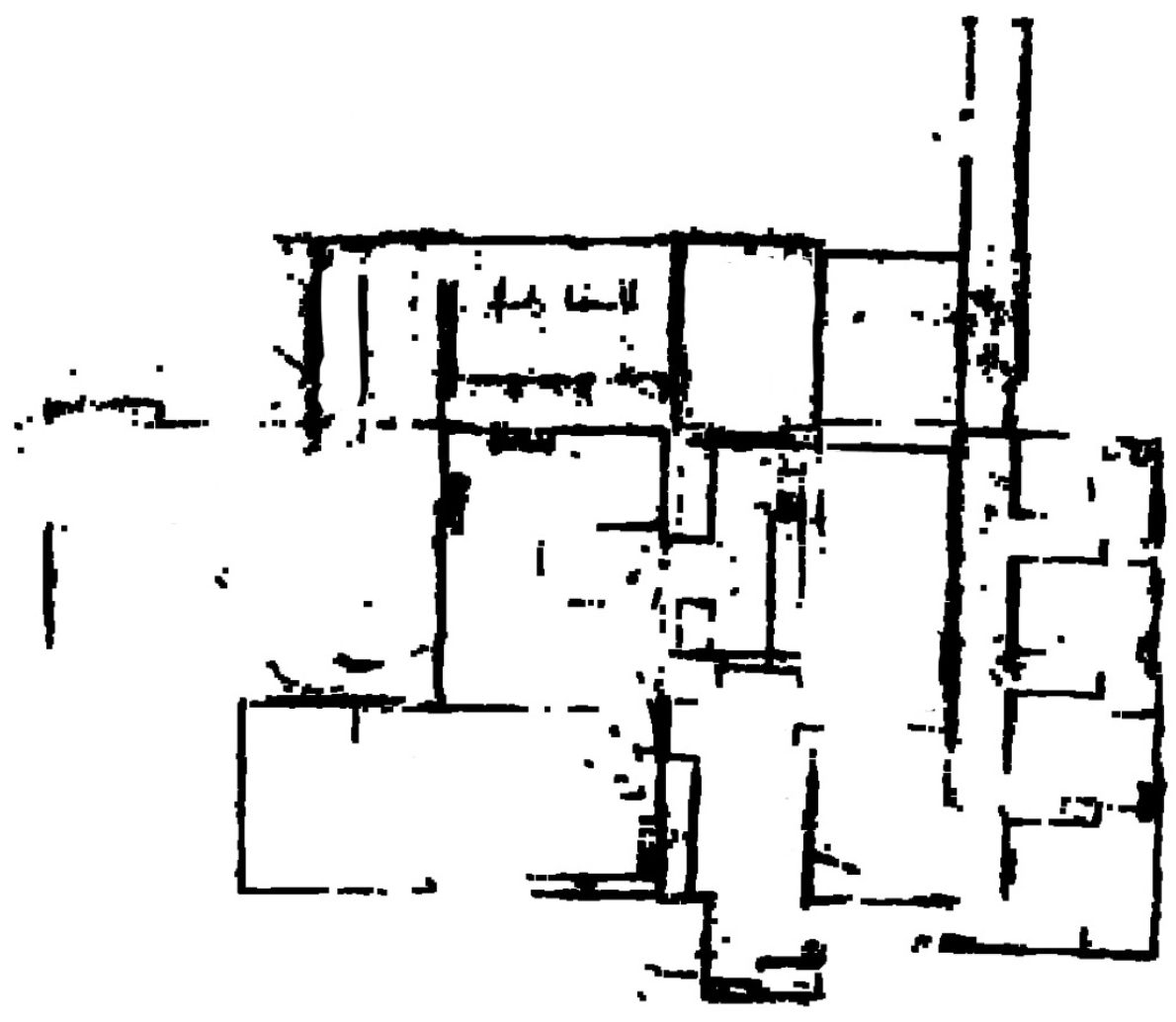}
        \label{fig:subfig4}
    }

    \caption{Comparison of using the proposed adaptive height threshold instead of different fixed height thresholds for the generation of top-down BEV map image from the 3D LiDAR map.}
    \label{fig:bev_thresh}
\end{figure}

\ph{Bird's-Eye View generation} Given a LiDAR pointcloud map generated by our SLAM method~\cite{reinke2022locus}, applying ortho-projection of map points on a 2D plane and point filtering using a fixed height threshold can be a fast technique to generate BEV images. However, an accurate threshold tuning is required to remove clutter and sensor noise (Fig. \ref{fig:bev_thresh}). A fixed threshold is not suitable for dynamic and unknown environments. For instance, it may fail in the case of a slanted environment or if objects are present in proximity of the ceiling (e.g. lights or chandeliers).

Motivated by the fact that walls are usually the densest part of a room, this work extends the approach in~\cite{mahmood2021learning} with an adaptive height-based threshold. In~\cite{mahmood2021learning} walls are extracted by projecting all 3D points on a 2D histogram and selecting only the highest frequency bins. However the density-based approach is not robust in case dense LiDAR points are acquired  from objects inside the room. For example, during testing, ceiling air vents can be observed in multiple pointclouds and thus result in false walls in the generated BEV images.
% The idea is therefore to extend this approach by introducing a height-based filter, with an adaptive threshold. 
By adding an adaptive height threshold to the density filter, with the intuition that walls are not only the densest parts but also have the greatest height difference between their lowest and highest points, the proposed approach is robust to ceiling hanging objects (Fig. \ref{fig:subfig4}). Thus, after initial filter to remove lowest floor points, the LiDAR map built up to time $t$ is voxelized first. The $x,y$ plane is then discretized into $1000\times1000$ bins, reflecting the pixel distribution of the BEV image, with each bin $B_i$, representing a column of points. The score $S_i$ is then computed as:
% , based on eq. \ref{equation:bin_score}.
%%%%%%%%%%%%%%%%%%%%%%%%%%%%%%%%%%%%%%%%%%%%%%%%%%%%%%%%%%%%%%
\begin{equation}
    S_i = [0.4 \cdot \sum_{p \in B_i}p] + [0.6 \cdot (\max_{Z} (p \in B_i) - \min_{Z}(p \in B_i))] 
    \label{equation:bin_score}
\end{equation}
%%%%%%%%%%%%%%%%%%%%%%%%%%%%%%%%%%%%%%%%%%%%%%%%%%%%%%%%%%%%%%
The first term, namely density score, is computed as the sum of points in the column (assuming equal surfaces). The second term computes the maximum height difference between points within the same bin. 
% The coefficients of the weighted sum give higher importance to the height gap, given its greater reliability in driving the distinction between clutter and structural information.
To account for varying height distribution within a cell and to adapt to the dynamic nature of the map, an adaptive height threshold is used. $T_{current}$ is computed by taking the mean of all bins scores and used to update the global threshold $T_t$ to generate the BEV image as: 
%%%%%%%%%%%%%%%%%%%%%%%%%%%%%%%%%%%%%%%%%%%%%%%%%%%%%%%%%%%%%%5
\begin{equation}
    T_t = 0.85 \cdot T_{t-1} + (1-0.85) \cdot T_{current}
    \label{equation:low_pass_thresh}
\end{equation}
%%%%%%%%%%%%%%%%%%%%%%%%%%%%%%%%%%%%%%%%%%%%%%%%%%%%%%%%%%%%%%
% The idea is to create a BEV image to capture the footprint of the environment, even in presence of noise and clutter. The BEV generator projects 3D map points onto a two-dimensional image plane. To deal with noise and clutter, a height filter is used. However, selecting the proper filtering threshold is challenging: as the cutting height increases, noise and clutter are gradually filtered out but it is possible to incur in a loss of structural information. To enhance the robustness of the process, an algorithm based on the height and density analysis of columns of voxels (explained in Section \ref{sec:exp}) is introduced. 
The output BEV image is finally generated to a $1000\times1000px$ image and with a padding of $12px$ per side.

\begin{figure}[t!]
    \centering
    \includegraphics[width=\columnwidth]{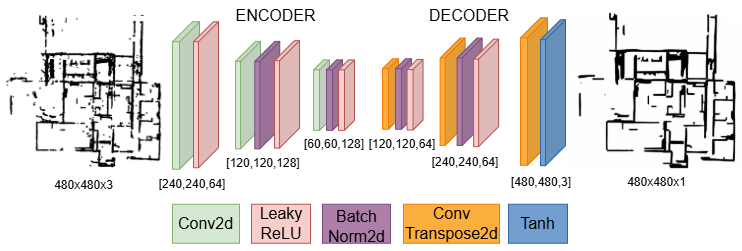}
    \caption{Architecture of the generator network of the proposed GAN: an encoder-decoder convolutional neural network extracting latent features from the input BEV image.}
    \label{fig:gan_net}
\end{figure}

\ph{GAN-based BEV denoising}
During noise removal actual structural information is often removed too, leading to an incorrect top-down image. For instance, gaps in walls incorrectly resembling connections between adjacent rooms might be generated. The same issue can also occur due to missing sensor measurements. To overcome this drawback, a GAN-based image inpainting~\cite{yeh2017semantic} method is utilized. The aim is to fill-in the missing (or lost) structural information and refine the quality of the overall image (e.g. closing false room connections). Beyond that, a GAN architecture has been chosen with the idea of having a tool also able to learn the data distribution for augmenting the training dataset with synthetic samples.
The generator model (Fig. \ref{fig:gan_net}) is an encoder-decoder convolutional neural network that is trained to take a BEV image with missing information as input  and generate a BEV image with reduced clutter noise and reconstructed walls. The training phase is driven by the loss function expressed in \ref{equation:g_loss}: %The initial dataset of BEV images is therefore augmented by using artificially corrupted salt and pepper noise to remove and add structural information, respectively, to the groundtruth images. 

%and train the GAN generator to restore the missing details. 
%%%%%%%%%%%%%%%%%%%%%%%%%%%%%%%%%%%%%%%%%%%%%%%%%%%%%%%%%%%%%%%%%%%%%%%
%%%%%%%%%%%%%%%%%%%%%%%%%%%%%%%%%%%%%%%%%%%%%%%%%%%%%%%%%%%%%%%%%%%%%%%
%The generator model (Fig. \ref{fig:gan_net}) is an encoder-decoder convolutional neural network extracting latent features from the input BEV image. It is trained on a modified loss function, expressed in \ref{equation:g_loss}: 
%%%%%%%%%%%%%%%%%%%%%%%%%%%%%%%%%%%%%%%%%%%%%%%%%%%%%%%%%%%%%%%%%%%%%%%
\begin{equation}
    L_G = 0.2 \cdot \min_{G} \log (1 - D(G(z))) + 0.8 \cdot L_1
    \label{equation:g_loss}
\end{equation}
%%%%%%%%%%%%%%%%%%%%%%%%%%%%%%%%%%%%%%%%%%%%%%%%%%%%%%%%%%%%%%%%%%%%%%%
The first term represents the discriminator's output computed on the restored input image. The $L_1$ loss encourages the generated images to be similar to the ground truth ones. %It is calculated as the mean of the absolute differences between the pixel values of the generated and groundtruth images.\looseness=-1

The discriminator model is a convolutional neural network composed of a sequence of 3 convolutional blocks, each followed by a BatchNorm and LeakyReLU activation function, and a final convolutional layer. The output is flattened and used to perform the binary classification among real or generated sample. The model is built on top of the PyTorch library and has been trained on GPU. The training set, consisting of the CubiCasa5K~\cite{kalervo2019cubicasa5k} images and samples collected in-field, has been augmented by using artificially corrupted salt and pepper noise to remove and add structural information, respectively, to the groundtruth images. During inference the model is deployed on CPU.

\ph{Segmentation engine}
The structural segmentation module processes BEV images to break down the complex environment structure into independent components. This block has been equipped to gather room-level information, but also integrates a pixel-to-voxel association technique to segment different places in the pointcloud map. This is done by projecting 2D masks back into the 3D domain (described in \ref{sec:scene_graph}). 

The core of the module is the Mask R-CNN \cite{he2017mask} segmentation model. 
%It is the state-of-the-art segmentation model, introduced as an extension of Faster R-CNN. 
Among the segmentation models that have achieved state-of-the-art performance, it is widely used due to its limited resource consumption, stability and ecosystem support. As an extension of Faster R-CNN, it introduces RoIAlign as a quantization-free layer to preserve exact spatial locations and adds a new branch for predicting segmentation masks on each Region of Interest (RoI) in parallel with the existing branch for bounding box recognition. The model, relying on the TensorFlow library, has been trained on GPU but a CPU-only version is used at runtime. %Moreover, mask and class prediction are decoupled to avoid competition among classes.
A custom dataset has been realized and used to train the model to perform such task. The dataset is an extension of the well known open-source floorplan dataset CubiCasa5K, merged with samples collected in-field and annotated by hand. Another collection of custom samples obtained via data augmentation has finally been added to the training set, with the aim of increasing its complexity. The input layer of the segmentation model, using a ResNet101 as backbone, has fixed a shape accepting $1024\times1024px$ BEV images. This technique allows different-sized maps to be represented in the same image space. After inference, the model produces instance masks in the same image coordinates. To maintain input and output pixel correspondences, the generated masked image is forced to have a $1024\times1024px$ resolution.

\subsection{Scene graph rendering}
\label{sec:scene_graph}
Once the information about the structure and the semantics of the environment is gathered, the last module in the pipeline aggregates all data to build the 3D hierarchical scene graph. To interactively monitor the outcome during operations, the graph is rendered inside the RViz visualization tool of the ROS ecosystem. This section describes in order how each layer is built.

\ph{Object layer} The first layer of the graph encodes position and object type about the artifacts spread in the area under investigation. MMDetection generates class ID and segmentation mask for each detected object, expressed in image coordinates. To retrieve artifacts' positions in world coordinates, a depth sensor (e.g. NeBula-Spot's Velodyne VLP-16 LiDAR) is needed, as well as a sensor fusion system to merge 3D and 2D data. 

The developed sensor fusion system computes the extrinsic parameters $E$ from the world reference frame, in which LiDAR points are expressed, to the optical frame of the camera sensor. Starting from $E$, world points are transformed in the camera coordinate system and finally projected on the image plane by using the camera intrinsic matrix $K$ (Eq. \ref{equation:homogeneous_camera}).
%%%%%%%%%%%%%%%%%%%%%%%%%%%%%%%%%%%%%%%%%%%%%%%%%%%%%%%%%%%%%%%%%%%%%%%%%%%%%%%%%%%%%%%%%%%%%
\begin{equation}
    P_{camera} = E_{world2optical} \times P_{world} \;, \; p_{image} = K \times P_{camera}
    \label{equation:homogeneous_camera}
\end{equation}
%%%%%%%%%%%%%%%%%%%%%%%%%%%%%%%%%%%%%%%%%%%%%%%%%%%%%%%%%%%%%%%%%%%%%%%%%%%%%%%%%%%%%%%%%%%%%
% \begin{equation}
%     \mathbf{p_{image}} = K \times P_{camera}
%     \label{equation:perspective_uv}
% \end{equation}
%%%%%%%%%%%%%%%%%%%%%%%%%%%%%%%%%%%%%%%%%%%%%%%%%%%%%%%%%%%%%%%%%%%%%%%%%%%%%%%%%%%%%%%%%%%%%
After extracting only the points falling into each object mask, these are reprojected back to the world coordinates to mark their positions. In this work, no depth estimation module has been employed but a heuristic approach based on bounding box sizes is used. A detection proximity module is finally incorporated to check that the same object is not represented multiple times in the scene graph.

\ph{Scene layer} 
The scene classification network is triggered periodically, whenever a sufficient amount of data is available to obtain a reliable classification. Each scene is marked in the second layer of the graph at the location of the robot when the classification is performed. Objects to $scene_i$ links are created by keeping track of artifact detections between the generation of $scene_i$ and $scene_{i+1}$. 

\ph{Room layer}
The segmentation module generates structural components' coordinates in the image domain. To retrieve the 3D spatial coordinates of each pixel, a pixel-to-voxel association pipeline is also introduced. Starting from the 3D LiDAR map, points are aggregated into voxels first. The whole map space is then discretized into a $1000 \times 1000$ grid representing the  resolution of the BEV image (without padding). Columns of voxels falling in the same grid cell are mapped to the corresponding pixel. For each room mask, the corners are then projected back to the 3D pointcloud and the centroid is finally used to identify its position in the scene graph.  
Scene to room connections are drawn by analyzing the spatial distribution of scene markers with respect to room layouts and arrangement.  

\ph{Building layer}
Rooms layout is finally used to abstract information up to building level. Adjacent rooms are supposed to belong to the same building while open spaces are used as the main discriminant factor to distinguish different buildings. 

\begin{figure}[t]
    \centering
    \includegraphics[width=0.9\linewidth]{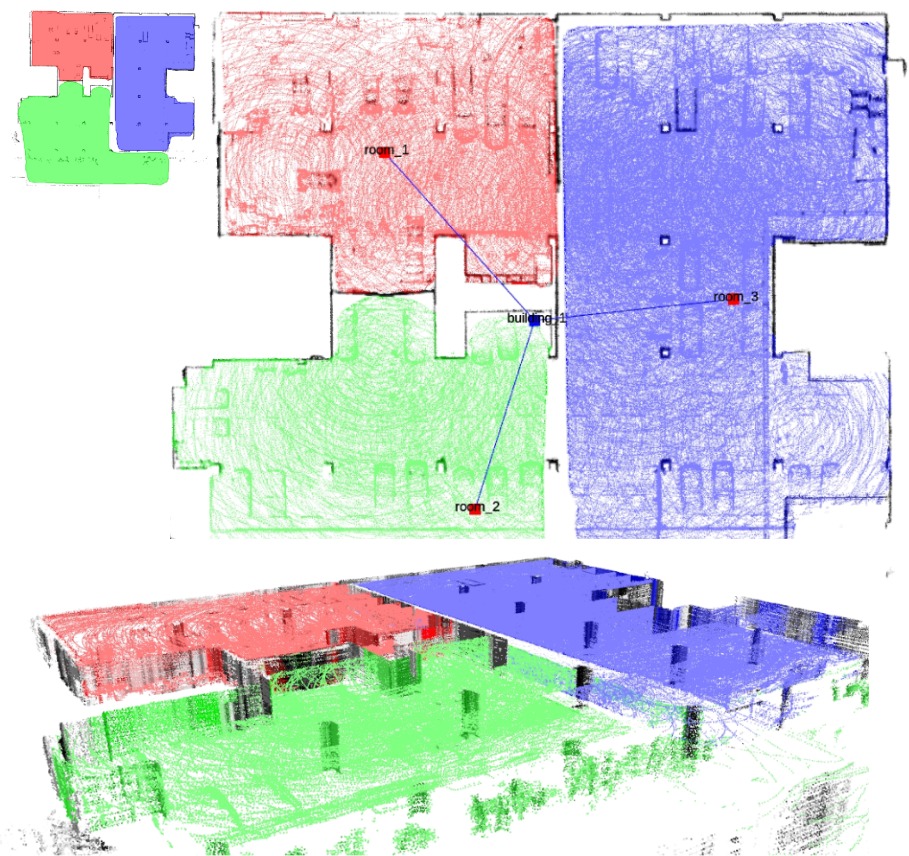}
    \caption{The pointcloud segmentation from 2D masks. The small image is the output of the structural segmentation process. The pixel-to-voxel projection pipeline retrieves the 3D points associated to each mask pixel in the image and sets the same color. Full video:~\url{https://youtu.be/KY3atFNdSts}}
    \label{fig:lamp_map}
\end{figure}

\ph{Segmented pointcloud map} 
To visualize BIM-like low-level geometry of the environment coupled with the high-level representation from the scene graph, the proposed framework integrates a fast pointcloud segmentation from 2D masks (Fig \ref{fig:lamp_map}). The procedure exploits the previously described pixel-to-voxel associations. For each pixel in the segmented image, all points falling inside the corresponding voxels are set to the same color. 

%%%%%%%%%%%%%%%%%%%%%%%%%%%%%%%%%%%%%%%%%%%%%%

%%%%%%%%%%%%%%%%%%%%%%%%%%%%%%%%%%%%%%%%%%%%%%
% Experimetns
\section{EXPERIMENTAL RESULTS}
\label{sec:exp}
The feasibility of Pix2G towards real-world applications was evaluated by conducing autonomous exploration experiments using a legged robot in complex indoor environments. The hardware setup of the robotic platform and the details of the testing areas in which experiments have been performed are listed in the following sections.
%The NASA JPL's NeBula-Spot~\cite{nebulaSpot} platform was deployed in two distinct environments. 
Quantitative and qualitative results are provided for all experiments.

\subsection{Experiment Setup}
\subsubsection{Robot and Sensors}
In each experiment, the autonomous NeBula-Spot~\cite{nebulaSpot} is deployed in an a-priori unknown environment. The robot is equipped with a VLP-16 LiDAR to provide pointclouds at 10Hz and an Intel Realsense D455 providing RGB-D data at 15Hz. No prior map is provided to the robot. It leverages the onboard LiDAR localization~\cite{fakoorian2022rose} and mapping~\cite{reinke2022locus} solution to build its own map and uses an exploration planner~\cite{moon2024efficient} to navigate in the environment. During in-field tests, the whole structural segmentation model is deployed on the robot's onboard CPU. Fig. \ref{fig:concept} shows an instance of the experiment setup.

\subsubsection{Environments} The purpose of real-world experiments is to show the generalization capabilities of the system and how Pix2G adapts to different scenarios. For this reason, two environments with diverse characteristics have been identified as testing areas. The first one is a \textbf{cluttered garage environment} spanning $300m^2$, with adjacent workshop-like space. This scene, featuring wide areas separated by temporary room dividers, challenges the segmentation model in open-space scenarios, where many approaches fail. On the contrary, the second experiment was conducted in an empty \textbf{urban office-like environment} showcasing long corridors with complex room structures. This scenario challenges the model with its narrow passages, cubicles and rooms with different sizes and shapes, for a total surface of $92m^2$.

\subsection{Results}

\subsubsection{Cluttered garage environment}
% In this experiment, the autonomous NeBula-Spot~\cite{nebulaSpot} robot explores a cluttered car garage with adjacent workshop-like space. No prior map is provided to the robot and utilizing an exploration planner~\cite{moon2024efficient} the robot incrementally builds the map of the environment using the onboard LiDAR localization~\cite{fakoorian2022rose} and mapping~\cite{reinke2022locus} solution. A VLP-16 LiDAR is utilized to provide pointclouds at 10Hz and an Intel Realsense D455 provides RGBD data at 15Hz. Figure \ref{fig:concept} shows an instance of this experiment.

For this experiment, Fig.~\ref{fig:architecture}, shows step-by-step the intermediate process to generate a 2D top-down BEV building information map and a 3D scene graph. The structural segmentation of the three large areas within this environment is highlighted in the top-left corner of Fig.~\ref{fig:lamp_map}. The successful result of the 2D structural segmentation is then utilized to colorize the 3D pointcloud for human understanding and to generate the build and room levels of the scene graph. The clear structural segmentation of the three different areas highlights the quality of the BEV image generation and GAN-based structural denoising components of the approach to generate a meaningful building information map. The accuracy of the structural segments can also be noted by the colorization of pointclouds, showing that in 3D space not overlapping points are miss-classified, hence resulting in an easily understandable 3D scene graph.

%Furthermore, to understand the applicability of Pix-to-graph for real-time operations, the compute requirements are evaluated in terms of execution time and memory usage. The performance is evaluated on a computer equipped with an Intel Core i7-9700K processor with  16 GB of RAM. All evaluations are performed on CPU to emulate the performance of constrained compute platforms.
% The system is also equipped with a NVIDIA GeForce RTX 4090 Graphic Processing Unit (GPU). To show that the proposed approach is meant to be run on constrained platforms, the GPU has been disabled and the whole stack is run on CPU. To stress each component, test have been performed in a large open floor environment (Fig \ref{fig:final_graph}).

Furthermore, to understand the applicability of Pix2G for real-time operations, the compute requirements are evaluated in terms of execution time and memory usage both on Base Station (BS) and onboard the robot. BS is equipped with an Intel Core i7-9700K processor with  16 GB of RAM, while the NeBula-Spot processing runs onto an Intel Core i7-8809G processor with 16 GB of RAM.
%%%%%%%%%%%%%%%%%%%%%%%%%%%%%%%%%%%%%%%%%%%%%%%%%%%%%%%%%%%%%%%%%%%%%%%%%%%%%%%
\begin{figure}[b!]
    \centering
    \includegraphics[width=0.95\columnwidth]{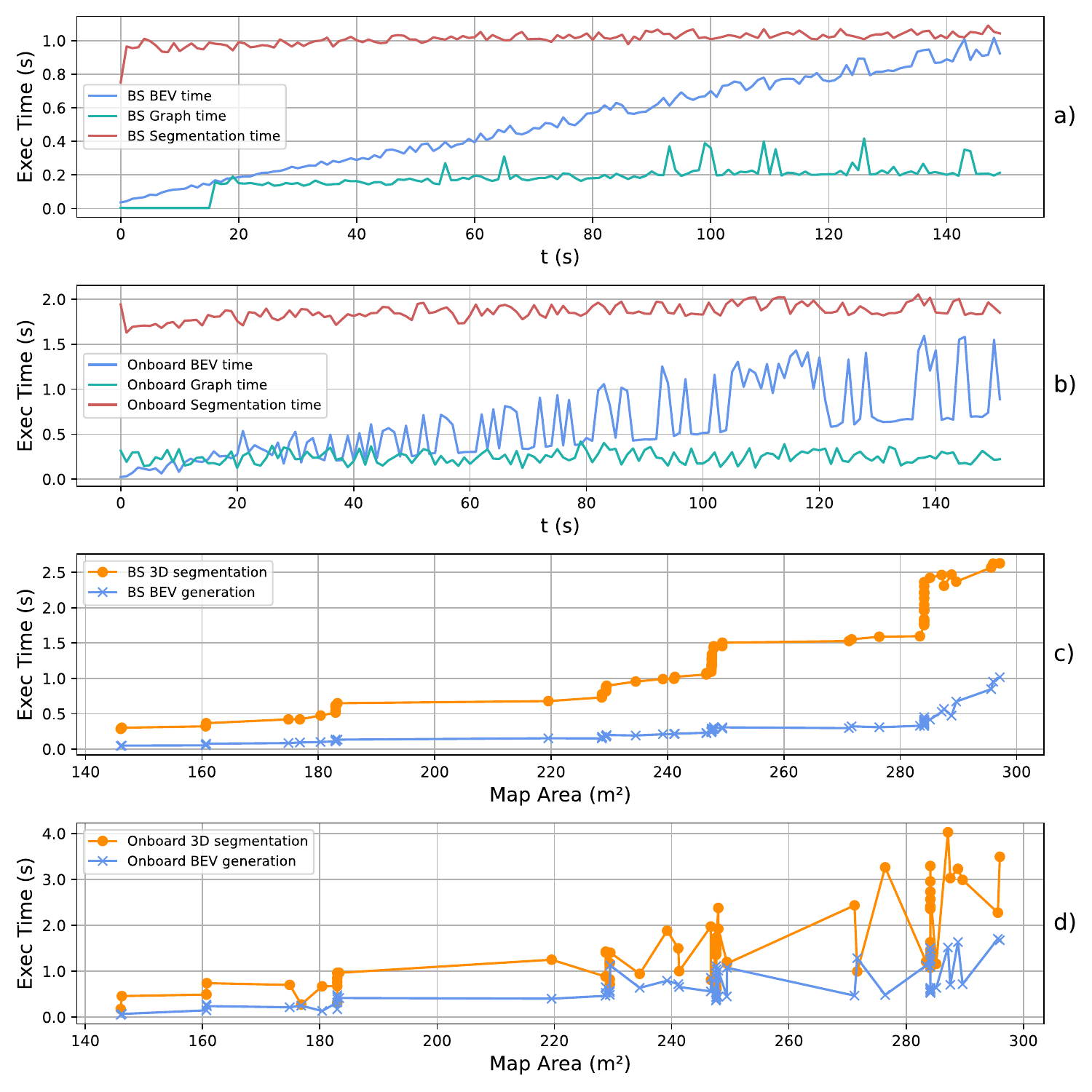}
    \caption{System performance results at runtime as executed on BS (a,c) and onboard the robot (b,d). While image segmentation and graph rendering time are almost constant, the time required to produce the BEV and construct the segmented pointcloud show a linearly growing trend based on the map size.}
    \label{fig:components_times}
\end{figure}
%%%%%%%%%%%%%%%%%%%%%%%%%%%%%%%%%%%%%%%%%%%%%%%%%%%%%%%%%%%%%%%%%%%%%%%%%%%%%%%

%%%%%%%%%%%%%%%%%%%%%%%%%%%%%%%%%%%%%%%%%%%%%%%%%%%%%%%%%%%%%%%%%%%%%%%%%%%%%%%
\begin{figure}[t!]
    \centering
    \includegraphics[width=\columnwidth]{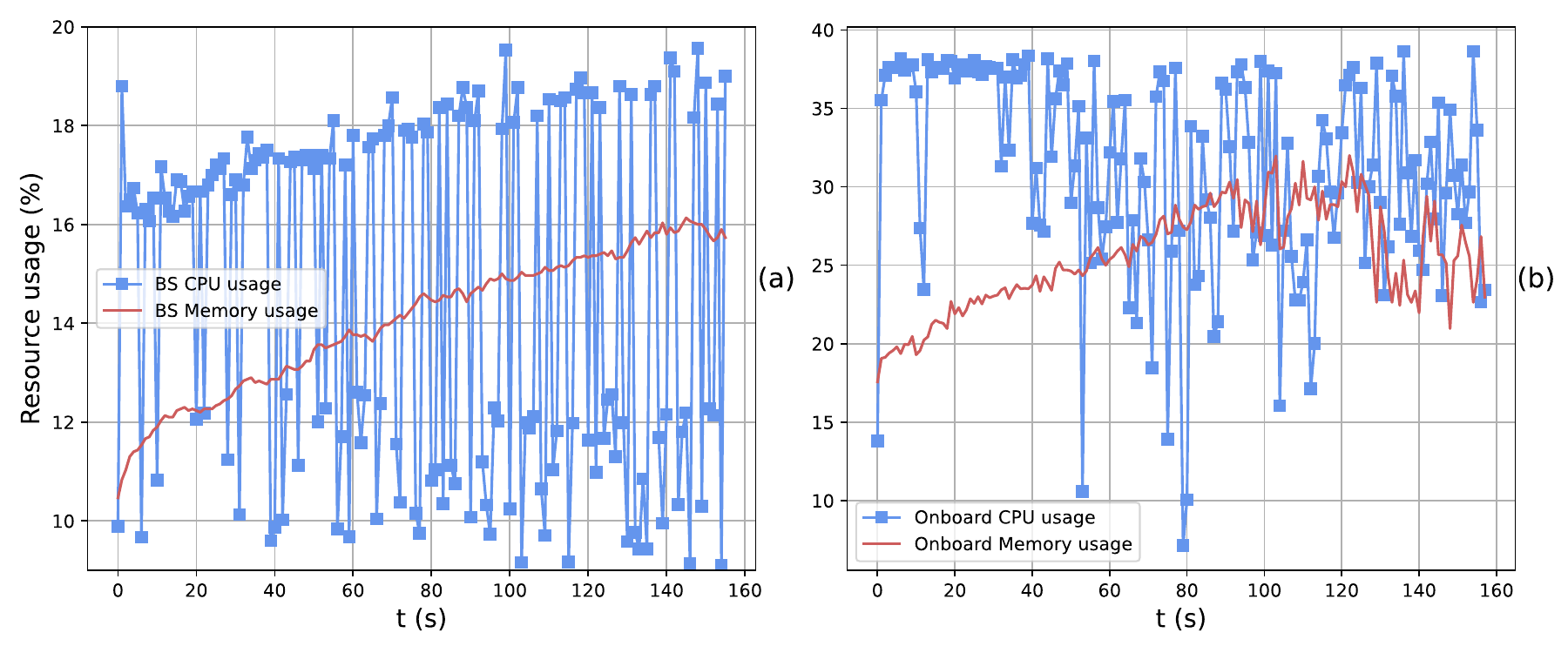}
    \caption{CPU and Memory usage of the proposed system as executed on BS (a) or onboard the robot (b).}
    \label{fig:res_usage}
\end{figure}
%%%%%%%%%%%%%%%%%%%%%%%%%%%%%%%%%%%%%%%%%%%%%%%%%%%%%%%%%%%%%%%%%%%%%%%%%%%%%%%
Figs.~\ref{fig:components_times}(a) and~\ref{fig:components_times}(b) indicate the execution times of each component when run on BS or onboard the robot, respectively. The segmentation inference and the graph rendering times are acceptable for an almost real-time approach. Nevertheless, the BEV generation showed a linear growth behavior over time. The BEV, as well as the 3D segmentation, depends on the size of the map. Charts \ref{fig:components_times}(c) and \ref{fig:components_times}(d) show how the execution of the 2 tasks varies at increasing map size. While the BEV still allows real-time operation, the adoption of a shifting window approach could be a future solution to bound these execution times. 
We finally evaluated the CPU and memory usage of Pix2G on BS, Fig.~\ref{fig:res_usage}(a), and on the robot, Fig.~\ref{fig:res_usage}(b). 
Peaks up to $\sim 40\%$ suggest that the framework is also suitable for onboard deployment. In communication degradation scenarios, this allows the robot to build the graph onboard and continue the mission in case communication to BS becomes unavailable. Moreover, running the framework locally allows reducing the robot-to-BS network traffic significantly.

%In Fig. \ref{fig:components_times}(a), the execution times of each single component are indicated. The segmentation inference and the graph rendering times are acceptable for an almost real-time approach. Nevertheless, the BEV generation showed a linear growth behavior over time. The BEV, as well as the 3D segmentation, depends on the size of the map. Chart \ref{fig:components_times}(b) shows how the execution of the 2 tasks varies at increasing map size. While the BEV still allows real-time operation, the adoption of a shifting window approach could be a future solution to bound these execution times. 
%We finally evaluated the CPU and memory usage of Pix-to-graph (Fig. \ref{fig:res_usage}). Peaks up to $\sim 20\%$ suggest that the framework is also suitable for constrained platforms.

\subsubsection{Urban office environment}
% To show the generalization capabilities of the proposed system, the second experiment is conducted in an empty urban office-like environment, characterized by long corridors, narrow passages, cubicles and rooms with different sizes and shapes. The experiment is conducted in the same settings as the garage, with no priors on the environment and the robot relying solely on its autonomy stack for navigation and exploration.\looseness=-1

\begin{figure}[b!]
    \centering
    \includegraphics[width=\linewidth]{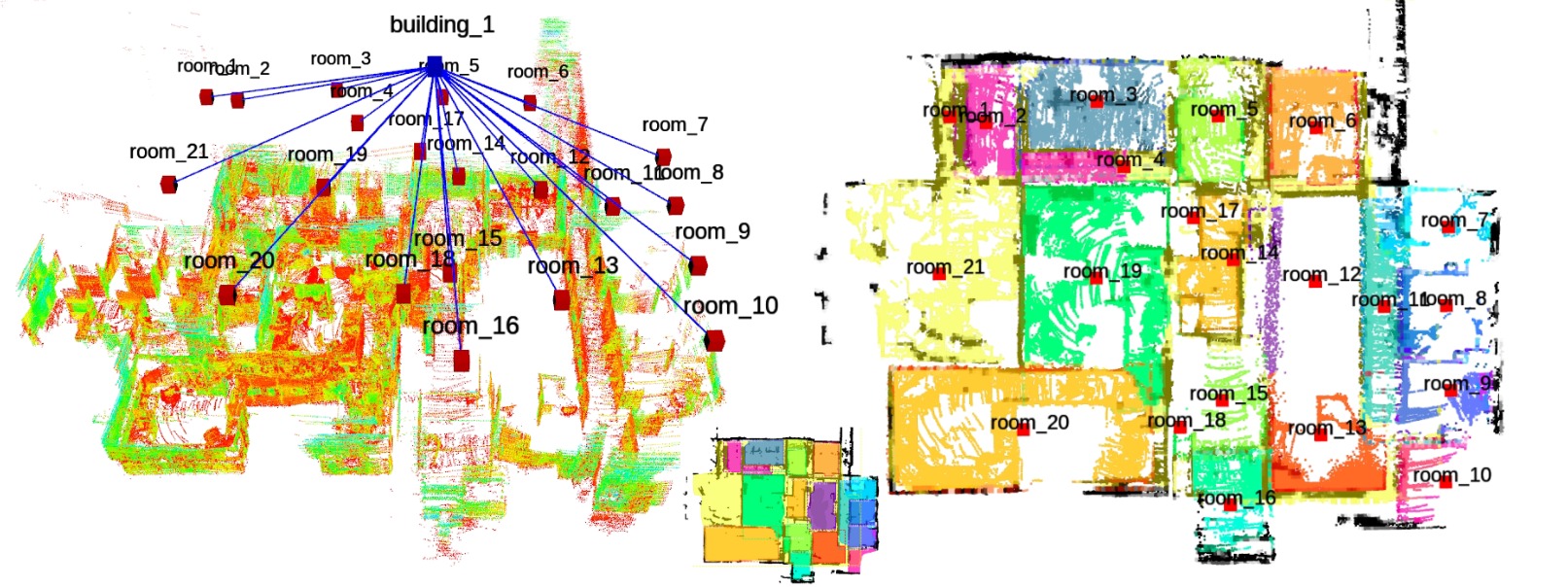}
    \caption{Output graph after a testing run in urban office environment. On the left is the original 3D map with room and building info, on the right the colorized map.\\Result video:~\url{https://youtu.be/w05V9aw4PeY}}
    % \url{https://youtu.be/w05V9aw4PeY}}
    \label{fig:final_b11_graph}
\end{figure}

In contrast to the few large spaces of the garage, in this experiment the NeBula-Spot robot moves across a large number of small adjacent rooms, divided by narrow corridors and small openings. Fig. \ref{fig:final_b11_graph} shows the accurate space partitioning that can be achieved in such challenging scenario thanks to Pix2G. The segmented BEV image in the middle shows how the introduction of adaptive filtering and GAN refinement makes the proposed system robust to incomplete acquisitions and map noise. The space partitioning system successfully identifies 21 different rooms in the explored environment. 
%%%%%%%%%%%%%%%%%%%%%%%%%%%%%%%%%%%%%%%%%%%%%%%%%%%%%%%%%%%%%%%
%%%%%%%%%%%%%%%%%%%%%%%%%%%%%%%%%%%%%%%%%%%%%%%%%%%%%%%%%%%%%%%

Results from real-world experiments show how the developed system robustly performs space partitioning and scene graph creation in challenging, a-priori unknown scenarios, in presence of map noise and clutter. The model performs well both in constrained environments and large open-spaces, where most of the approaches in the literature fail.

\subsubsection{Ablation Study}

We finally evaluate the efficacy of the proposed training method for the structural segmentation task. 3 case studies are analyzed. The first version of the Mask R-CNN model, namely study $A$, has been trained on BEV images collected from testing runs inside the JPL campus and annotated by hand. None of the training samples was taken from the two testing areas considered in the final experiments. To account for generalization, in study $B$, the training set has been extended with the CubiCasa5K \cite{kalervo2019cubicasa5k} dataset. It is a large-scale image dataset, containing $\sim 5000$ samples annotated into over 80 floorplan object categories, aiming at overcoming the lack of representative datasets to investigate the problem of automatic floorplan images parsing. One of the limitations of using CubiCasa5K as-is in this work, is that it contains only clean and well-formed floorplan images. To account for noise in BEV images, in study $C$, the dataset has been extended with additional samples, collected by the CubiCasa5K and augmented with random salt and pepper noise at varying percentages and sizes. In this study, the GAN preprocessing is also introduced in the pipeline.

In the experiment, the validation set comprising dirty CubiCasa5K samples has been adopted as test-bench. The mean Intersection over Union (mIoU), measuring the overlapping of each predicted mask with respect to the ground truth one, Precision (P) and Recall (R) have been adopted as evaluation metrics.

\begin{table}[h!]
\setlength{\tabcolsep}{3pt} % Adjusts the column spacing
\begin{tabular}{ >{\centering\arraybackslash}p{0.18\columnwidth}  >{\centering\arraybackslash}p{0.23\columnwidth} 
>{\centering\arraybackslash}p{0.23\columnwidth} 
>{\centering\arraybackslash}p{0.24\columnwidth} }
\hline
 & 
\textbf{mIoU \%} & 
\textbf{P \%} & 
\textbf{R \%}\\
\hline
Study A & 
54.92 &
62.51 &
58.34 \\

Study B & 
84.36 &
87.84 &
86.53 \\

Study C & 
92.73 &
92.41 &
90.68 \\
\hline
\end{tabular}
\caption{Performance comparison of the proposed models in the structural segmentation task.}
\label{tab:performance_results}
\end{table}

Results in table \ref{tab:performance_results} show how the use of GAN-based noise filtering and the introduction of CubiCasa5K dataset lead to a considerable performance enhancement. The mIoU increases as a consequence of improved segmentation masks. Higher Precision indicates that the predicted instances are actually correct, while a lower number of missed detections leads to an increase in the Recall score.

\begin{figure}[b!]
  \centering
  \subfigure[]{  
    \includegraphics[width=0.45\linewidth]{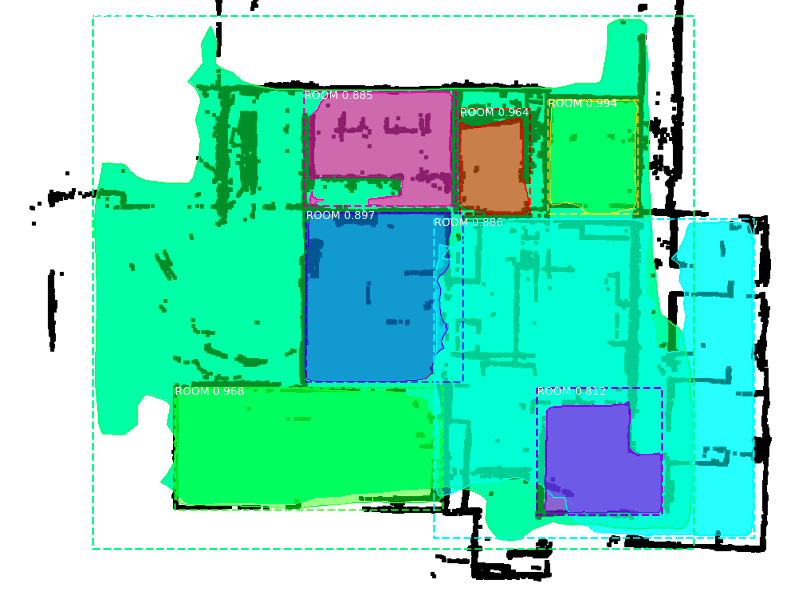}
    \label{fig:comp_a}
  }
  \hfill
  \subfigure[]{
  \includegraphics[width=0.45\linewidth]{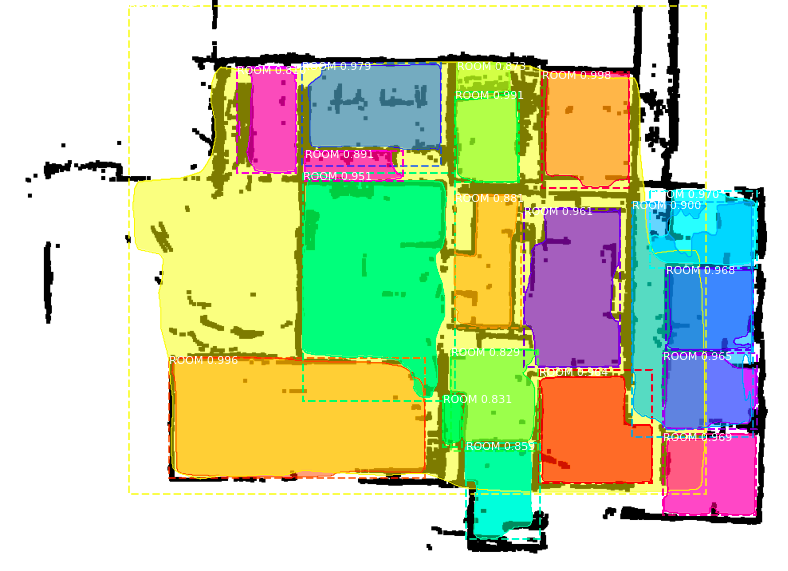}
  \label{fig:comp_b}
  }
    \caption{Comparison between the model trained on the initial dataset comprising only training samples from JPL spaces and the one trained on augmented CubiCasa5K.}
    \label{fig:struct_comparison}
\end{figure}

Fig. \ref{fig:struct_comparison} provides a qualitative demonstration of the enhancement in the structural segmentation performance achieved by Mask R-CNN trained on the final augmented dataset, compared to the initial one comprising JPL images only.

\section{CONCLUSIONS AND FUTURE WORK}
\label{sec:conclusions}
We presented Pix2G, a novel lightweight framework to generate structured scene graphs from raw pixels during autonomous robotic exploration of unknown environments. The output of the method is a multi-layer graph representing information at 4 levels of abstraction: object level, scene level, room level and building level. The framework leverages Mask R-CNN as segmentation core, trained on an augmented version of the CubiCasa5K dataset. To account for sensor noise and clutter, an adaptive filtering system with a GAN architecture for image inpainting are also introduced.   
We demonstrated the capabilities of the proposed framework on a set of real-world data collected during autonomous explorations of the NeBula-Spot robot, both in open and narrow unknown environments. A performance analysis of the structural segmentation system on real-world and synthetic data is also provided. Future works will investigate how the proposed framework scales to multi-level scenarios and multi-robot settings. Solutions to bound the BEV computational burden with growing maps will be investigated, as well as segmentation approaches in presence of dynamic clutter. Future research will also explore how the generated global semantic-rich representation can be integrated into recent Vision-Language-Action (VLA) paradigms.
%%%%%%%%%%%%%%%%%%%%%%%%%%%%%%%%%%%%%%%%%%%%%%

%                                   % the textheight of the last page by a suitable amount.
%                                   % This command does not take effect until the next page
%                                   % so it should come on the page before the last. Make
%                                   % sure that you do not shorten the textheight too much.

\bibliographystyle{IEEEtran}
\bibliography{bibliography}

\end{document}